\documentclass[sigconf]{acmart}
\newcommand{\bt}{}

\newcommand{\ft}{}

\usepackage[per-mode=symbol]{siunitx}
\usepackage{lscape}
\usepackage{pifont}
\usepackage{xcolor}
\usepackage{makecell}
\usepackage{soul}

\newcommand{\cmark}{\textcolor{green!80!black}{\ding{51}}}
\newcommand{\xmark}{\textcolor{red}{\ding{55}}}

\AtBeginDocument{%
  \providecommand\BibTeX{{%
    \normalfont B\kern-0.5em{\scshape i\kern-0.25em b}\kern-0.8em\TeX}}}

\acmConference[CHI '24]{Proceedings of the CHI Conference on Human Factors in Computing Systems}{May 11--16, 2024}{Honolulu, HI, USA}
\begin{document}

\title{Towards Robotic Companions: Understanding Handler-Guide Dog Interactions for Informed Guide Dog Robot Design}

\author{Hochul Hwang}
\email{hochulhwang@cs.umass.edu}
\orcid{0000-0002-3199-7208} 
\affiliation{%
  \institution{University of Massachusetts Amherst}
  \city{Amherst}
  \state{Massachusetts}
  \country{USA}
}

\author{Hee-Tae Jung}
\email{heetjung@iu.edu}
\orcid{0000-0001-8921-570X} 
\affiliation{%
  \institution{Indiana University Indianapolis}
  \city{Indianapolis}
  \state{Indiana}
  \country{USA}
}

\author{Nicholas A Giudice}
\email{nicholas.giudice@maine.edu}
\orcid{0000-0002-7640-0428} 
\affiliation{%
  \institution{University of Maine}
  \city{Orono}
  \state{Maine}
  \country{USA}
}

\author{Joydeep Biswas}
\email{joydeepb@cs.utexas.edu}
\orcid{0000-0002-1211-1731} 
\affiliation{%
  \institution{University of Texas at Austin}
  \city{Austin}
  \state{Texas}
  \country{USA}
}

\author{Sunghoon Ivan Lee}
\authornote{Co-corresponding authors.}
\email{silee@cs.umass.edu}
\orcid{0000-0001-5935-125X} 
\affiliation{%
  \institution{University of Massachusetts Amherst}
  \city{Amherst}
  \state{Massachusetts}
  \country{USA}
}

\author{Donghyun Kim}
\authornotemark[1]
\email{donghyunkim@umass.edu}
\orcid{0000-0001-9534-5383} 
\affiliation{%
  \institution{University of Massachusetts Amherst}
  \city{Amherst}
  \state{Massachusetts}
  \country{USA}
}


\renewcommand{\shortauthors}{Hwang, et al.}

\begin{abstract}
  Dog guides are favored by blind and low-vision (BLV) individuals for their ability to enhance independence and confidence by reducing safety concerns and increasing navigation efficiency compared to traditional mobility aids. However, only \bt{a relatively small proportion} of BLV individuals work with dog guides due to their limited availability and associated maintenance responsibilities. There is considerable recent interest in addressing this challenge by developing legged guide dog robots. This study was designed to determine critical aspects of the handler-guide dog interaction and better understand handler needs to inform guide dog robot development. We conducted semi-structured interviews and observation sessions with 23 dog guide handlers and 5 trainers. Thematic analysis revealed critical limitations in guide dog work, desired personalization in handler-guide dog interaction, and important perspectives on future guide dog robots. Grounded on these findings, we discuss pivotal design insights for guide dog robots aimed for adoption within the BLV community. 
\end{abstract}





\maketitle

\section{Introduction}
Around the world, \bt{there are over 250 million blind and low-vision (BLV) people. This number is expected to rise to more than 700 million by 2050, largely due to the increasing and aging population~\cite{ackland2017world}. Notably, within the United States, this demographic is expected to double, mirroring the worldwide trend~\cite{varma2016visual}}. Among the various assistive tools supporting safe and efficient navigation, a critical aspect of daily life, dog guides stand out as remarkably effective companions, enhancing mobility, confidence, and independence for these individuals~\cite{miner2001experience, whitmarsh2005benefits, wiggett2008experience}. For example, formally-trained dog guides are known to enable handlers to be more mobile~\cite{refson1999health,lloyd2008guide2}, walk faster~\cite{clark1986efficiency}, navigate more efficiently by taking direct routes instead of following edges or shorelines~\cite{giudice2008blind}, and exhibit superior travel performance, with indicators including orientation, mobility, and difficulty with travel~\cite{lloyd2008guide}, compared to people who use \ft{a} long-cane. However, only a small fraction of BLV people benefit from dog guides~\cite{GD101, CANADA}. The main factors underlying this are the limited availability of dogs and the substantial cost and time investment required by guide dog schools to train and deploy a dog guide (approximately \$40,000 USD and a 2-year training period\bt{~\cite{pit23,geb23,cnc23}}). Moreover, not all trained dogs ultimately serve as dog guides~\cite{menuge2021early}, further underscoring the scarcity of these invaluable resources.

Even in cases where a dog guide is potentially available for placement with a BLV handler, the decision to work with a dog guide is a complex \ft{and} multifaceted process. Some BLV people may not meet health requirements for obtaining a dog guide (e.g., the ability to walk several miles a day or suffer from dog allergies)~\cite{GDoA, GDF}, while others might lack the financial means to support the essential care and maintenance costs \bt{(approximately \$1,000 USD each year~\cite{pit23})}. And those who choose to work with a dog guide also take on the responsibilities of ownership, including exercise, feeding, and grooming, which can be burdensome for the handler. Furthermore, individuals might have to confront the challenge of managing emotional consequences from the retirement of a dog guide~\cite{schneider2005practice}. 

In contrast to their animal counterparts, robots offer unparalleled mass productivity, cost-effectiveness \bt{(commercially available quadruped robot starting from less than \$2,000 USD~\cite{ur23})}, low maintenance, and longevity, paving the way for innovative mobility aids. The utilization of robotic systems to enhance mobility for BLV individuals traces back to as early as 1976 (i.e., the MELDOG project)~\cite{tachi1978study, tachi1984guide}. Since then, researchers have proposed various forms of robotic guides, including smart canes~\cite{wahab2011smart, slade2021multimodal, zhang2023follower}, suitcase- or cart-shaped~\cite{guerreiro2019cabot, zhang2023follower, kuribayashi2023pathfinder}, and drones~\cite{avila2017dronenavigator, al2016exploring, tan2021flying}. More recently, owing to recent advancements in quadruped robots~\cite{kim2019highly, hwangbo2019learning}, four-legged robotic systems have gained attention as a potential solution to enhance the mobility of BLV people~\cite{hamed2019hierarchical, xiao2021robotic, chen2022can, chen2023quadruped, due2023guide, hwang2023system, kim2023transforming, liu2023dragon}. Quadruped robots offer natural and efficient travel mechanisms that enable navigation through uneven terrains and complex structures in real-world environments, such as stairs and curbs, that BLV people encounter in their daily lives. \bt{However, our research highlights a critical gap in guide dog robot studies: insufficient attention to user needs and requirements. This oversight can be largely attributed to the absence of user-centered research within robotics that examines the interactions between dog guides and their handlers in real-life scenarios. Such research is essential to discern which qualities of \ft{dog guides} should be preserved in robotic counterparts and to identify specific challenges that need to be addressed to overcome the limitations inherent in \ft{dog guides}.}

To \bt{identify such qualities and} bridge \bt{the} knowledge gap, we conducted semi-structured interviews with 23 dog guide handlers and 5 professional dog guide trainers. Our aim was to gain a comprehensive user-driven understanding of the interaction dynamics between handlers and their dog guides \bt{to identify the key qualities that should be translated to guide dog robots}. Based on these data, we extracted valuable insights to inform the development of a guide dog robot. A subset of seven dog guide handlers were also engaged in observation sessions, during which we observed handler-dog guide units navigating familiar surroundings. The qualitative data resulting from these interactions was subjected to rigorous thematic analysis, yielding meaningful codes and patterns that were further consolidated into overarching themes.

Through our multi-pronged qualitative studies, we discovered a compelling need motivating the development of a new mobility aid in the form of a quadruped robot (or the \textit{guide dog robot})~\cite{hersh2010robotic}. This revelation stems from anecdotal evidence extracted from the rich data obtained from interviews that underscore the inherent limitations within current handler-guide dog interactions, which include unidirectional communication between the handler and the \ft{dog guide}, the \ft{dog guide’s} limited ability to understand situationally important contextual information, and the \ft{dog guide}’s lack of flexibility in adapting to new environments and confusing situations. Subsequently, we delve into the continuous interactions between handlers and their dog guides which is necessary for personalizing the \ft{dog guide}'s behaviors to fit the handler’s needs. We recognize the value of these evolving interactions, as they stand as a foundational aspect for the design of effective robotic mobility aids but have not yet been systematically investigated and documented in the existing literature. Lastly, we explore the expectations and concerns among dog guide handlers in the context of guide dog robots based on their lived experiences. Drawing from these findings, we present new design implications for future guide dog robots. This paper sheds light on the discrepancies between real-world handler-guide dog interactions and previous guide dog robot development efforts, emphasizing the importance of incorporating human-centered design principles for BLV people. By highlighting recent advancements achieved in robotics, we aim to draw more awareness and attention to designing and developing a new and highly effective user-validated robotic mobility aid as an additional option for BLV individuals, aligning with the growing interest within the HCI community~\cite{mack2021we, brule2020review}. 

In summary, this paper contributes to HCI and robotics research on guide dog robot development for BLV people as follows:

\begin{itemize}
    \item We compile knowledge established by dog guide schools and professional trainers to characterize the mismatch between handler-guide dog interactions and emerging guide dog robot development.
    \item Based on semi-structured interviews and observation sessions conducted with dog guide handlers and trainers, we emphasize the demand and utility for a new mobility aid by revealing the limitations in the handler-guide dog interaction.
    \item We present the continuous interaction between the handler and \ft{dog guide} and highlight the key components that are personalized to the handler's needs that may be pivotal for the real-world deployment of the robot.
    \item We propose, to the best of our knowledge, the first set of user-driven guide dog robot design guidelines in terms of hardware, software, and interface to foster guide dog robot development.
\end{itemize}

Given the prevailing norm in language usage, namely \textit{guide dog}, we will use this terminology instead of the formal \textit{dog guide} throughout the rest of the paper.

\section{Background} 
\label{sec:background}

This section provides background information on standardized approaches used by handlers and guide dogs to interact and cooperate in facilitating nonvisual navigation. The information presented in this section, developed over the past century in the field of guide dog training~\cite{igdf}, forms the foundational knowledge upon which our findings are built. We will begin by dissecting the fundamental components of nonvisual navigation: orientation and mobility. We will then detail the formation of the handler-guide dog partnership, drawing upon seminal literature and official documents from guide dog schools.

\begin{figure*}
  \includegraphics[width=0.85\textwidth]{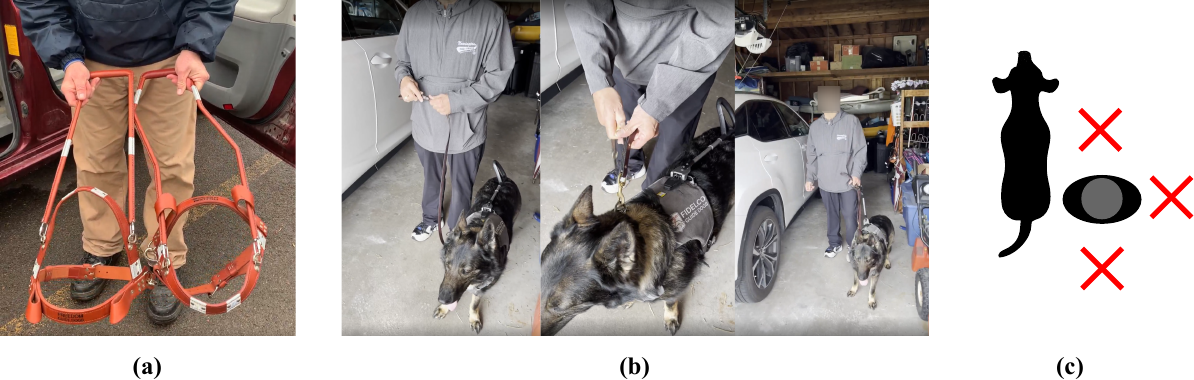}
  \caption{Connection through the harness. (a) Various types of harness handles exist to match different guide dog handlers. For example, the offset harness on the right can be used for handlers who tend to step on their guide dog's foot due to their gait. (b) The handler is transitioning from grabbing an extended leash to grabbing both the leash and the harness handle before walking. (c) The handler is trained to be positioned on the right side of the guide dog while walking, specifically next to the hindquarters, often grabbing the leash and the harness handle with the left hand.}
  \label{fig:harness}
  \Description{Figure 1: "Title: Connection through the harness. Column (a): Person holding two harness handles – one straight, one offset. Column (b): Sequential images of a guide dog handler adjusting the leash and grabbing the harness handle for walking. Column (c): Top-view diagram of handler-guide dog unit. "X" marks signify incorrect walking positions (front, right, and back of handler)."}
\end{figure*}

\subsection{Shared Responsibility in Orientation and Mobility Among Handlers and Guide Dogs}
\label{subsec:shared}
\textit{Orientation} and \textit{mobility} are the foundational concepts that define the collaborative effort between handlers and guide dogs in facilitating safe and effective navigation~\cite{wiener2010foundations}. Orientation entails knowledge of one's position and facing direction in relation to landmarks and keeping track of these spatial relationships, while mobility is the act of safely and efficiently moving through space. The combination of these two skills facilitates purposeful and directed movement within the surrounding environment, resulting in effective navigation. In a typical navigation scenario from point A to point B, the environment may feature guiding landmarks (e.g., curbs, poles, shops) and hindrances (e.g., construction cones, pedestrians, holes). Orientation enables individuals to monitor their spatial orientation relative to these landmarks, while mobility facilitates secure, unobstructed progress between these landmarks and ultimately toward the destination~\cite{wiener2010foundations, giudice2008blind}. 

In the partnership between guide dogs and their handlers, a shared responsibility model emerges, with each playing distinct but interdependent roles~\cite{Tucker84}. Handlers take on a pivotal role in \textit{orientation} while guide dogs assume a vital role in \textit{mobility}. In other words, handlers are responsible for the general orientation to the surroundings and making navigational decisions to reach the destination by giving “cues” (i.e., commands) to their guide dogs. Guide dogs, on the other hand, bear the responsibility of ensuring safe mobility by responding to the cues provided by their handlers. While supporting mobility, they are trained to maintain a straight line and avoid obstacles and potential hazards in the path of travel~\cite{Tucker84, wiener2010foundations}. Guide dogs also possess the capacity to “intelligently disobey” when they perceive that following the cue would pose a significant risk to the handler. For instance, they may refuse to proceed if there is a fast-approaching moving object that could endanger the handler. This shared strategy ensures effective navigation while leveraging the respective strengths of both the handler, primarily responsible for orientation and the guide dog, concerned with safe mobility.
    
\subsection{Formation of the Handler-Guide Dog Unit}
\subsubsection{Matching}
The process of pairing guide dogs with handlers constitutes a pivotal step in establishing a proficient handler-guide dog partnership. This matching process involves a comprehensive assessment of both the handler and the guide dog. “Class-ready” guide dogs, having successfully completed the required evaluation outlined in Appendix~\ref{sec:evaluation}, are paired with potential guide dog handlers based on various factors. The criteria for matching, while varying among different guide dog schools, encompassed factors, such as the potential handler's age, height, weight, cause of blindness, location (city and state), daily routine, walking speed, and pulling force~\cite{bane2020}. 

\subsubsection{Connection and positioning}
The handler-guide dog unit is connected mainly through a rigid harness~\cite{Tucker84}, although various handle designs exist to accommodate variations in the physical attributes of both the handler and the guide dog, including height differences and handler preferences, as depicted in Figure~\ref{fig:harness}a. While handlers often hold a leash and the harness handle simultaneously, the leash serves specific purposes, such as boarding buses, navigating stairs or escalators, walking alongside a human sighted guide, corrections, or relieving the guide dog, as illustrated in Figure \ref{fig:harness}b. However, the rigid harness primarily serves as the connection during walking, offering precise control and enhanced safety. Importantly, its rigid design is critical for enabling immediate stops when necessary due to its direct connection. The handle's angle is also crucial, with a recommended 30-degree inclination to allow effective tracking of the guide dog's movements~\cite{Tucker84}.

In general, the handler is positioned on the right side with respect to the guide dog, using the left hand to hold onto the harness. This enables the handler to search or carry objects with their right hand. Specifically, the correct position of the handler is beside the guide dog’s hindquarters during walking. This proximity enables the handler to take an extra step and stand next to the guide dog's shoulders when the guide dog halts~\cite{Tucker84}, as depicted in Figure \ref{fig:harness}c. This positioning ensures safety by preventing the guide dog from cutting in front of the handler and facilitates various interactions, including efficient turns and maneuvering behavior.

\subsubsection{Handler-Guide Dog Interaction \bt{and Prior Studies}}
Effective communication between the handler and guide dog is essential for clearly delineating their roles and responsibilities (e.g., orientation and mobility) to ensure safe and efficient navigation. The handler gives verbal cues, hand signals, and combinations of them to the guide dog to convey intentions while navigating. Particularly, handlers employ verbal cues for basic obedience, controlling speed and direction, as well as directional indicators. In response, the formally trained guide dog complies or intelligently disobeys based on the situation. The handler then perceives the guide dog’s motions that are transmitted to the handler’s hand through the rigid handle~\cite{wiener2010foundations}. Maintaining consistent pulling strength or tightness of the harness is crucial \bt{for both handler satisfaction~\cite{craigon2017she} and to ensure clear communication of direction to the guide dog\bt{~\cite{hwang2023system,bane2020}}}.

\bt{The existing body of qualitative research offers a range of insights into the handler-guide dog interaction, yet it lacks extensive and systematic surveys essential for the development of effective mobility aids. Wong~\cite{wong2006perceptions} highlighted the predominantly positive experience reported by guide dog handlers, noting exceptions, such as unwanted attention towards the dog and mobility constraints without their dog. Similarly, a study by Miner~\cite{miner2001experience} interviewed eight guide dog handlers and reported their affirmative experiences and insights on mobility while working with a guide dog. Rickly et al.~\cite{rickly2021}, on the other hand, reported some challenges encountered by guide dog handlers, particularly related to transportation accessibility, encompassing issues like service refusal and access. Gaunet and Besse~\cite{gaunet2019guide} assessed the behavior of 23 guide dogs and blindfolded trainers in four different wayfinding navigation tasks. Our study extends this body of research by identifying key attributes of handler-guide dog interaction that should be retained in designing a guide dog robot.}

\subsubsection{Needs for a User-Driven Guide Dog Robot Guideline}
\bt{Existing research has provided valuable guidelines for user-driven robotic guides, focusing largely on non-legged systems, addressing the required technologies and challenges involved in developing such robotic aids~\cite{hong2022development, hersh2010robotic}. Others have adopted a participatory design approach to develop robotic guides focusing on indoor settings~\cite{azenkot2016enabling,feng2015designing}. Additionally, Zeng et al.'s study on handle interfaces, which involved testing a haptic handle interface in guiding robots with blindfolded individuals, has contributed to the understanding of user interfaces in robotic guides~\cite{zeng2018hapticrein}. However, these studies often do not adequately capture the detailed, real-life interactions between handlers and guide dogs, an aspect that may be critical for the development of legged guide dog robots.}

Building upon recent legged locomotion advancements, researchers have embarked on \ft{developing} guide dog robots to \bt{assist BLV people, leveraging the natural efficiency of legged robots in navigating complex terrains, such as stairs and curbs, that BLV people frequently encounter. This initiative, supported by studies by Hamed et al.\cite{hamed2019hierarchical}, Xiao et al.\cite{xiao2021robotic}, and others~\cite{chen2022can, chen2023quadruped, hwang2023system, kim2023train, due2023guide, defazio2023seeing}, represents a collaborative effort in the field. Xiao et al.~\cite{xiao2021robotic} propose a local path planner by formulating an optimization problem considering the two modes (taut or slack) of the human-leash-robot relationship. Wang et al.~\cite{chen2022can} conducted an indoor and outdoor walking experiment with BLV people to compare a quadruped and a wheeled robot. They showed that BLV participants preferred the wheeled robot over its quadruped counterpart (Cyberdog by Xiaomi~\cite{Xiaomi}) due to the quadruped’s rough motion and noise while walking. Hauser et al.~\cite{hauser2023s} evaluated pedestrian experiences (or humans’ perceptions of robots) during human-robot encounters under different configurations (e.g., no user-robot connection, leash connection, and harness connection) using a Spot robot (Boston Dynamics\cite{BostonDynamics}).}

\bt{Although there are growing efforts to adapt quadruped robots for BLV people, current user-centered research has not comprehensively captured key aspects necessary for the development of guide dog robots. Moreover, most quadruped robots designed for BLV individuals have failed to sufficiently meet user needs and requirements. For instance, some studies used a soft leash~\cite{xiao2021robotic} or an elastic rope~\cite{chen2023quadruped}, as the primary connection method, which is inappropriate due to safety issues. Moreover, certain studies position the handler far behind the robot, which is not ideal for high-speed walking and safety due to the increased range of motion while turning and the delay in response time, confirmed by guide dog trainers. Therefore, we aim to identify essential qualities of guide dogs that should be mirrored in robots and to pinpoint specific challenges to address the limitations of animal guide dogs, contributing to the field by steering towards a more user-centric approach in guide dog robot development, ensuring their design and functionality align closely with the needs and experiences of BLV individuals.}

\section{Method}
\label{sec:method}

Our study seeks to improve our comprehension of the practical interactions between handlers and guide dogs. Specifically, we aimed to investigate (1) biological guide dogs’ capabilities and fundamental limitations, (2) how handlers interact with their guide dogs in real-world settings, including whether they adapt or modify the recommended interaction methods to suit their individual needs, and (3) the expectations and concerns of guide dog handlers regarding guide dog robots based on their experiences. We hypothesized that the obtained information would offer invaluable insights to inform the development of guide dog robots. To accomplish this, we conducted semi-structured interviews and observation sessions involving guide dog handlers and trainers. In this section, we detail our study design and data analysis methods, with approval obtained from the University of Massachusetts Amherst IRB.

\begin{table*}
  \caption{Guide Dog Handler Demographics and Collected Data}
  \label{tab:gdhs}
  \begin{tabular}{ccccccc}
    \toprule
    \textbf{Participant} & \textbf{Age} & \textbf{Gender} & \textbf{Vision Level} & \textbf{\makecell[c]{Experience\\(years / \# of GDs$^{\dag}$)}} & \textbf{Interview} & \textbf{Observation} \\
    \midrule
    H01 & --  & F & Totally blind & 36 / 4 & \cmark & \cmark \\
    H02 & 69 & M & Totally blind & 30 / - & \cmark & \cmark \\
    H03 & 63 & M & Totally blind & 11 / 2 & \cmark & \cmark \\
    H04 & 41 & M & Totally blind & 20 / 4 & \cmark & \ft{--} \\
    H05 & 69 & F & Totally blind & 36 / 5 & \cmark & \ft{--} \\
    H06 & 57 & F & Totally blind & 34 / 4 & \cmark & \ft{--} \\
    H07 & 77 & M & Totally blind & 9 / 1 & \cmark & \ft{--} \\
    H08 & 66 & F & Totally blind & 0.5 / 1 & \cmark & \ft{--} \\
    H09 & 69 & M & Totally blind & 53 / 8 &  \cmark & \cmark\\
    H10 & 67 & F & Totally blind & 43 / 6 &  \cmark & \ft{--}\\
    H11 & 36 & F & Totally blind & 16 / 3 &  \cmark & \ft{--}\\
    H12 & 69 & F & Legally blind & 30 / 4 &  \cmark & \cmark\\
    H13 & 59 & M & Totally blind & 21 / 3 &  \cmark & \cmark\\
    H14 & 70 & M & Totally blind & 17 / 3 &  \cmark & \ft{--} \\
    H15 & 81 & M & Totally blind & 14 / 4 &  \cmark & \ft{--}\\
    H16 & 70 & F & Totally blind & 7 / 1 &  \cmark & \ft{--} \\
    H17 & 75 & F & Totally blind & 51 / 6 &  \cmark & \ft{--} \\
    H18 & 77 & F & Totally blind & 34 / 7 &  \cmark & \ft{--} \\
    H19 & 78 & F & Totally blind & 31$^{*}$ / 3 &  \cmark & \ft{--} \\
    H20 & 70 & M & Totally blind & 33 / 6&  \cmark & \ft{--} \\
    H21 & 64 & F & Totally blind & 6 / 1 &  \cmark & \cmark\\
    H22 & 60 & F & Totally blind & 16 / 4 &  \cmark & \ft{--} \\
    H23 & 25 & F & Totally blind & 7 / 1 &  \cmark & \ft{--} \\
    \bottomrule
\end{tabular}
\Description{Table 1: "Guide Dog Handler Demographics and Collected Data. This table contains information for 23 guide dog handlers, with each row representing an individual participant. The columns include subject ID, age, gender, vision level, experience (years/number of guide dogs), and participation in interview and observation sessions. Each row corresponds to a unique participant, providing their respective details in the columns mentioned above."}
\footnotesize{\\All subjects are legally blind having various conditions - H05, H06, H14: some light perception, \ft{H08: 20/2800 in one eye,} H12: 20/200 in right, 20/400 in left (no peripheral vision), H13: 20/2400, can see shapes, but no color distinction, H21: can see shapes, 20/2200 in one eye; $^{\dag}$ Indicates the number of guide dogs the participant worked with; $^{*}$ The subject did not work with a guide dog at the time of the interview.}\\
\end{table*}

\begin{table*}
  \caption{Guide Dog Trainer Demographics and Collected Data}
  \label{tab:gdts}
  \begin{tabular}{cccccc}
    \toprule
    \textbf{Participant} & \textbf{Age} & \textbf{Gender} & \textbf{\makecell[c]{Experience\\ (years)}} & \textbf{Interview} & \textbf{Observation} \\
    \midrule
    T01 & 65 & M & 45 & \cmark & \ft{--} \\
    T02 & 52 & M & 19 & \cmark & \cmark \\
    T03 & 58 & M & 35 & \cmark & \cmark \\
    T04 & 41 & M & 22 & \cmark & \ft{--} \\
    T05 & 34 & F & 9 & \cmark & \ft{--} \\
    \bottomrule
\end{tabular}
\Description{Table 2: "Guide Dog Trainer Demographics and Collected Data. This table contains information for 5 guide dog trainers, with each row representing an individual participant. The columns include subject ID, age, gender, experience (years), and participation in interview and observation sessions. Each row corresponds to a unique participant, providing their respective details in the columns mentioned above."}
\footnotesize{\\ The affiliations of the guide dog trainers with three guide dog schools were omitted from the table \\due to their potential for participant identification and perceived lack of significance.}\\
\end{table*}

\subsection{Participants}
\bt{We recruited handlers and trainers through snowball sampling from December 2021 through August 2023. We initially contacted organizations of BLV individuals and guide dog handlers~\cite{bscb20, gdui23}, three visually impaired service centers~\cite{carroll23}, and 13 guide dog training schools via emails and phone calls. The emails contained an IRB-approved flyer and a video that was later audio-described for enhanced accessibility. While the majority of participants were recruited primarily in the United States, one participant from the Republic of Korea was recruited through the snowball sampling \ft{process}.} \bt{The recruitment of the seven participants for the observation session was based on several considerations. These included the availability of the participants who met our inclusion and exclusion criteria, the distance the researchers needed to travel to their local community, the complexity of their environment, and the handlers' capability to navigate within their local community with a guide dog without the need for human assistance.}

The inclusion criteria for the guide dog handlers required them to (1) be 18 years or older, (2) have visual acuity equivalent to or less than legal blindness defined in the Social Security Act \S 1614~\cite{ssa20}, and (3) have at least six months of experience working with a guide dog. For guide dog trainers, the inclusion criteria were to (1) be 18 years or older and (2) have five or longer years of experience in guide dog training. Participants were compensated $\$50$ for the interview and an additional $\$50$ for observation sessions, except for one participant who refused compensation. Demographics of the guide dog handlers and trainers are detailed in Table~\ref{tab:gdhs} and Table~\ref{tab:gdts}, respectively.

All participants signed (or verbally agreed \ft{to}) an approved consent form. As a result, 23 handlers and 5 trainers were recruited. The handlers’ age median is 69 years and the lower (Q1) and upper quartiles (Q3) of the data are 60.75 years and 70 years respectively, having three outliers. The median of their duration working with guide dogs is 21 years and the lower and upper quartiles of the data are 11 years and 34 years respectively. Most of the handlers introduced themselves as totally blind although some participants provided more detail of their visual acuity (see Table~\ref{tab:gdhs}). 

\subsection{Procedure}
We conducted observation sessions and semi-structured interviews with guide dog handlers and professional guide dog trainers from three different guide dog training organizations, which are all accredited by the International Guide Dog Federation. The interview questions were developed and refined with the help of a guide dog handler and a trainer. The questions focused on guide dog functionalities and handler-guide dog unit interactions. The interview was conducted either via Zoom or in person with masks, in line with COVID precautions until early 2022.

In order to achieve an in-depth understanding of the interactions between the handler-guide dog unit in their natural environment and to gain hands-on experience with the guidance of professionals, we incorporated observation sessions during the interviews. \bt{These sessions involved observing guide dog handlers as they navigated familiar environments with their \ft{guide }dogs, offering real-life demonstrations of their interactions. Additionally, \ft{observation} sessions with guide dog trainers offered insights into the training processes of the handler-guide dog unit.}

\subsection{Data Collection}
We gathered comprehensive data, encompassing demographic information, over $1,940$ minutes of audio-recorded interviews, and $240$ minutes of video-recorded observation sessions (excluding redundant viewpoints). Subsequently, all recorded data were transcribed into text. One interview conducted in Korean was transcribed from daglo~\cite{dag23}, then translated from Korean to English using Naver Papago~\cite{pap23}, and further modified for grammatical soundness. \bt{Additionally, non-verbal interactions between guide dogs and handlers observed during observation sessions were transcribed into text for analysis.}

\subsection{Data Analysis}
The data were analyzed using thematic analysis~\cite{braun2006using}. The first, second, and corresponding authors individually analyzed a subset of the transcriptions using an open coding scheme. Specifically, Microsoft Word and Google Docs comments were utilized for open coding to create codes. Then, all authors contributed to generating affinity maps (i.e., axial coding) using the Miro software by connecting codes. Iterative processes of open coding and discussion continued until all authors reached the saturation of codes and agreed on the coding scheme. Initially, the first, second, and corresponding authors individually coded approximately five interviews each to create an intermediate coding scheme. Subsequently, the first author coded the rest of the interview \bt{and observation} data according to the coding scheme. Then, all the authors reviewed the codes and developed the themes in an iterative manner until all authors came to an agreement on the themes. We clearly state that generative AI technologies using large language models, notably ChatGPT, were not used for data analysis.

\section{Findings}
\label{sec:findings}
Our study reveals three pivotal findings: 1) crucial limitations within the handler-guide dog interactions, which may open opportunities for an additional robotic mobility aid, 2) the continuous interactions personalized over time and tailored to align with the handler’s specific needs, and 3) the perspectives, sentiments, and challenges encountered by guide dog handlers in the context of guide dog robots. 

\subsection{Critical Limitations of Interactions with Guide Dogs}
The limitations discovered within the handler-guide dog interaction, such as the guide dogs lacking human-level environmental awareness, unidirectional communication between handler and guide dog, and inflexibility in adapting to unfamiliar environments and change, provide insights into the potential design implications and open opportunities for guide dog robots.

\subsubsection{Challenges in Achieving Human-Level Environmental Awareness}
From our interviews, we uncovered a crucial theme: the struggle of guide dogs to comprehend the context of their surroundings as humans. Although guide dogs must grasp the nuances of the encountered situations and their surroundings, it is challenging for these animals to fully understand as their ability to have human-level environmental awareness of their surroundings is limited by their inherent intelligence.

The inability to understand the environmental context often confounded guide dogs to find important navigational objects or landmarks. For example, in the bustling cityscape, finding an empty seat on public transportation can be a triumph or a blunder for the handler-guide dog unit. Several BLV handlers highlighted the significance of this seemingly mundane task of finding an empty seat, whether in public transportation or in a doctor's office. Indeed, among the 17 guide dog handlers we discussed about finding empty seats, only seven were confident in their ability to locate these vacant seats with their guide dog. The majority found themselves inadvertently settling into an already occupied seat. Teaching guide dogs to identify an empty seat might seem straightforward, but due to the lack of understanding of the surrounding context, guide dogs often fail to identify a seat as H22 mentioned, “But they don't always know that the end table is not a seat. They just know it's a place that's low enough and you can sit on.” We found that this was not just about seats. H07's guide dog struggled to locate a door attached to glass walls and the handler's car, despite regular interactions with both. 

Another crucial aspect that guide dogs lack is a comprehensive understanding of spatial constraints for obstacle avoidance. All guide dog schools that participated in the GDUI survey~\cite{gdui23} train their guide dogs to detect, indicate, or avoid overhead obstacles, such as hazardous tree branches and signs, underscoring their recognition of the potential threat posed by such obstacles. Out of 17 BLV individuals, two expressed confidence in their ability to avoid overhead obstacles. The majority recounted unfortunate incidents involving collisions with tree limbs, branches, construction signs, tapes, scaffoldings, and even the side mirrors of trucks. \bt{This difficulty is consistent with findings from previous studies~\cite{williams2013pray,jeamwatthanachai2019indoor}, which have frequently highlighted it as a limitation.}

Teaching guide dogs to recognize overhead obstacles pose challenges, as these animal companions are naturally inclined to focus on monitoring ground\ft{-level} rather than \ft{chest-level (or above) spatial extent}. Moreover, we found that the nature of the overhead obstacle adds complexity to the problem. T05 shared that the objects' weight and density come into play, with lighter objects like tree branches proving harder to detect compared to the more conspicuous obstacles like poles. These challenges are further compounded by the way branches naturally grow and how they bend due to weather conditions like rain or snow. \bt{Although some handlers expressed a reluctance to walk with their guide dogs in the rain, they occasionally need to commute during a drizzle or encounter unexpected showers, }as H05 confirmed:

\begin{quote}
\textit{“My first dog, I remember [...] It was always raining or snowing or sleeting or something when I got him, and there were these branches and he was going so fast it knocked the hood off my head.”} - \textbf{H05}
\end{quote}

H10's story added a twist, recounting the struggles of navigating rope barriers in \ft{the queue at} a bank, where the guide dog couldn't grasp the need to navigate around them, as going under was a straightforward option for the dog. 

\subsubsection{Unidirectional Communication}
\label{subsec:unidirectional}
One critical limitation that handlers perceive when working with biological guide dogs is the inability to receive essential information related to unpredictable or accidental situations encountered during navigation. While the handler can convey navigational instructions to the guide dog through gestures or verbal cues, the communication from the guide dog to the handler is restricted to haptic feedback via the rigid harness, as outlined in Section~\ref{sec:background}. This constrained one-way communication often falls short of conveying vital details about the surrounding environment, potential hazards, or the risk of collisions, sometimes resulting in tragic accidents. Throughout our interviews, a recurring theme emerged: the difficulties stemming from the absence of verbal feedback from guide dogs.

When guide dogs stop for potential dangers to ensure safety, the responsibility to identify the risk and the reason for the stop falls on the handler to discern, as H14 mentioned, “they stop and let you try [to] figure out what is in front of you, why they're stopping. They make you figure it out. They can't communicate why, but they just stop”. In such situations, guide dog handlers resort to tactile exploration, using their foot or hand to identify potential sources of danger (H03 and H11). When handlers cannot find a clear reason, they may decide to move forward, which can lead to accidents, ranging from relatively moderate incidents like tripping off a curb or falling into a pool, as H01 and H03 mentioned, to more severe adversities, including the loss of life. T01 highlighted the case of a guide dog handler who had the ability to travel often but made a wrong decision due to a misperception of the environment which could have been prevented by giving appropriate feedback to the handler.

\begin{quote}
\textit{“One client I'm talking about, she died falling in the subway. She didn't have a dog from me anymore and it wasn't the dog's fault, because she let go of the dog as she fell in the subway. She told the dog to get on a train that wasn't there. Multi platform subway station. She heard the train went over, but she swore it was her train.”} - \textbf{T01}
\end{quote}

In fact, we found that handlers often desire to receive more explanation from their guide dog about the environment and situation. The majority of the interviewed handlers expressed the wish to receive information about the surrounding environment from the guide dog, such as details about upcoming corners and streets or the name of the store they are passing by in a shopping center. H11 who resides in an urban area wished to know if there were any upcoming construction, as she mentioned, “In terms of like practical things, I guess if they could tell me about oh, we're coming up to some construction or things like that, that might be nice too”. H12, who mainly used a bus before retirement, highlighted that it would be helpful to know if it is the right bus to be getting on. H06 expressed a desire to obtain information about traffic light colors, especially in noisy intersections where it is difficult to interpret the traffic flow by listening. H14 emphasized the value of any communicable information: “Anything that can be communicated will be useful. Anything. Change in the sidewalk, change in the road. Some cars are doing that now. They'll tell you the difference in the pavement or a difference in the... whatever. Anything to acclimate you to your situation.” Regarding indoor settings, H03 and H22 noted that having information about the people in a meeting room, dimensions, and furnishings, such as available seats, would be immensely beneficial. 

\subsubsection{Low Adaptability to Unfamiliar Environments and Changes}
\label{subsec:inflexibility}
Guide dogs achieve their full potential through extensive training and experience. However, rapid adaptation to unfamiliar environments or situations is not their strong suit. Nevertheless, handlers \ft{inevitably need to} visit new locations, and our living spaces and travel environments \ft{are} continuously \ft{evolving. This includes} changes in traffic rules, object placement, environment layouts, and technology, \ft{all of} which collectively present ongoing challenges. Our research shows that handlers and trainers find it difficult to provide additional training programs, aids, or assistive devices to compensate for a guide dog's limited adaptability in the face of unfamiliar or evolving environments. 

When guide dog handlers visit unfamiliar places, they often rely on sighted guides or GPS-based devices and apps for navigation assistance. However, sighted guides may not always be accessible, and independent GPS devices may fall short of fulfilling handlers’ requirements. H06 and H09, for instance, pointed out that they need to simultaneously use multiple apps depending on the situation, which is inconvenient and quickly drains both battery and data. Others used visual interpreting service apps, such as Aira~\cite{Aira-app}, which connects to a sighted guide located remotely to provide navigational assistance using the smartphone’s camera. However, as H06 described, obtaining reliable and confident support from a remote guide can sometimes be challenging. H11 also emphasized the inconvenience of using such an app when carrying items in their hands. H04, a South Korean urban resident, expressed that using GPS-based apps like Google Maps is difficult in a densely populated, high-rise environment due to the GPS sensing inaccuracy, a concern echoed by H17, a New York City resident.

Adapting to evolving environments also presents unforeseen complexities in guide dog work, posing a significant challenge. Long-term guide dog handlers, with over 40 years of experience, highlighted the increased difficulty of travel compared to earlier times with only parallel traffic. Factors such as the Right-Turn-on-Red principle, crossing islands, blended sidewalks and streets, and the proliferation of electric cars were identified by H10 and H17 as contributing to the heightened complexity of safe and efficient navigation in urban settings. T02 noted that while guide dogs can detect nearby low-speed electric cars, those approaching from a distance or at high speed still pose a challenge:

\begin{quote}
    {“The biggest challenge now with traffic is all the new hybrid car and electric car because they're don't make any noise, they very silence and clients have a hard time judge traffic or know if there is a car coming or not. If it's really very close to or turning or something, then the dog will, like intelligent disobedience, the dog will stop. But if it's coming from far away or it's coming fast or something, the dogs will not really judge this correctly.”} - \textbf{T02}
\end{quote}

\subsection{Continuous Interactions are Required to Adjust the Behaviors of Animal Guide Dogs to Meet the Handlers' Needs}
\label{subsec:personalization}
Comprehensive information regarding the interaction methods between guide dogs and their handlers, the range of services provided by guide dogs, and the training mechanisms is crucial when developing a guide dog robot. This information is readily available in the published literature, as we have summarized in Section~\ref{sec:background} and Appendix~\ref{sec:apx-gduisurvey} - Table~\ref{tab:gduisurvey}. However, we also discovered that the interaction between the handler and the guide dog does not always strictly adhere to the established methods. In some cases, handlers may deviate from the prescribed method taught by the training schools to align with their specific needs. For example, we discovered that interactions between handlers and guide dogs evolve over time, driven by day-to-day guiding work and additional training provided by the handlers. This signifies that class-ready guide dogs when initially matched with a handler, may not be performing at their full capacity. It is in the period following their pairing and collaborative work with the handler as a team that guide dogs truly grasp and adapt to each handler’s unique requirements, ultimately refining their behavior to become a “well-patterned” (e.g., personalized) guide dog. We delve into the evolving nature of these interactions, shedding light on what guide dogs autonomously learn and what handlers teach their dogs to emphasize the importance of personalization. This emphasis could provide valuable insights into the development of an effective guide dog robot.

\subsubsection{Learning to Autonomously Navigate Routes}
As \ft{is} detailed in Section~\ref{sec:background}, the handler-guide dog unit’s roles are distinct: handlers oversee orientation, while guide dogs take charge of managing mobility. However, over time, these roles may become somewhat blurred, particularly in environments that the unit regularly visits. With increasing guide dog’s familiarity along a route, guide dogs often overtake the role of orientation. Consequently, the frequency of interactions between the handler and the dog substantially decreases over time. For instance, the handlers may skip directional cues such as “turn left” or “turn right” on over-learned routes or locations, as H11 mentions, “I just have always found that it's just really by repetition and by ... I don't really have to have a specific command. It's more by doing it enough and she learns it that way.” 

As this learning process continues to evolve, guide dogs may even navigate to frequently visited places autonomously conditioned on that the route is relatively simple and short. We observed this phenomenon during our observation session with H01, who stated, “I'm not giving him a direction. I'm just hanging on to the [harness]." as the participant walked toward the office door where she works. H16 also shared a similar experience of reaching a destination without any cues:

\begin{quote}
    {“I can take a walk of my brother's house is about four miles away in \ft{Winchester}. And once, we cross the street about a block from my house, he knows where we're going and I really don't have to say anything the rest of the way. I do. But I don't really have to. He just takes me right to their back door.”} - \textbf{H16}
\end{quote}

However, it is important to emphasize that even with extended familiarity, handlers maintain ultimate responsibility for orientation and retain control over the overall navigation task. A guide dog trainer emphasized this point, stating that

\begin{quote}
    {“The dogs, even though the dogs aren't responsible for orientation, they start to pattern pretty well on regular routes. [However] The person is responsible for knowing where they are [...]”} - \textbf{T04}
\end{quote}

The autonomous orientation learned from experience by guide dogs can occasionally lead to inconvenience. This can manifest when a dog assumes a route to a frequently visited location when the actual destination lies beyond that familiar spot. For instance, H10 and H12 shared experiences where their guide dogs directed them to a regularly visited grocery store while they were on route to a different place. In both cases, they had to intervene and redirect their guide dogs by firmly saying “No” when the dogs made incorrect autonomous decisions. However, it is noteworthy that handlers often refrain from correcting this behavior to maintain consistency in their guide dogs' navigation habits. This reluctance stems from the desire to preserve the convenience of the developed autonomy for frequent visits to familiar places, as T03 pointed out:

\begin{quote}
    {“But you don't really correct the dog because some days you want the dog to go there.”} - \textbf{T03}
\end{quote}

Our findings collectively suggest that simply transferring the orientation role from a handler to an autonomous navigation algorithm in robots may not be an appropriate approach for guide dog robot development. Identifying formal criteria to assess the appropriate level of autonomy transfer is an open question. However, our findings clearly reveal that subtle variations in navigation roles exist within the unit, and handlers rely on or correct the autonomy of their guide dogs depending on the situations and goals at hand.

\begin{figure*}
  \includegraphics[width=.9\textwidth]{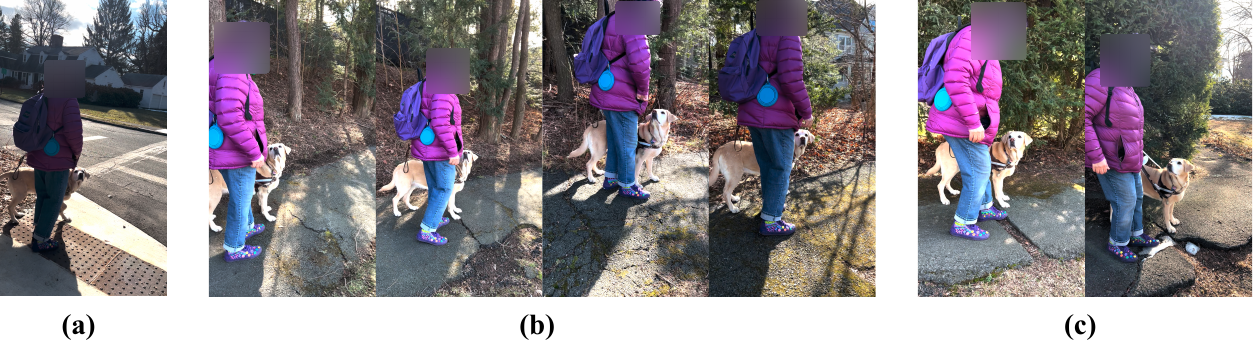}
  \caption{Observation of personalized stops. (a) The guide dog is trained to stop at tactile pavings from the guide dog school. (b) H21 trained her guide dog to stop at cracks in the sidewalk. (c) H21 also trained her guide dog to stop at larger cracks existing in H21's local community.}
  \label{fig:stop}
  \Description{Figure 2: "Title: Observation of personalized stops. Column (a): Handler-guide dog unit stopped at the tactile paving. Column (b): Handler-guide dog unit stopped at cracks in the sidewalk. Column (c): Handler-guide dog unit stopped at larger cracks in the sidewalk."}
\end{figure*}

\subsubsection{Additional Training to Teach New Behaviors}
One specific training that handlers prioritize is teaching new verbal commands, landmarks, vocabulary, and behaviors, all of which are tailored to the handlers’ individual needs and their unique living environments. Indeed, underscoring its significance,  “home training” involves trainers personally conducting handler-guide dog unit training in the vicinity of the handler's home. To accomplish these customized objectives, handlers often utilize a standardized training technique called back-chaining (also referred to as tagging, targeting, and patterning). H18 articulated by stating, 

\begin{quote}
    {“The basic commands are for all the guide dogs. And then I hone in my guide dog's training to the specifics that I need. [...] and now it's up to me to reinforce the things that he [guide dog trainer] has taught us, dog and me. And then I will be increasing her training to specifics that I need in my life."}- \textbf{H18}
\end{quote}

For instance, T03 and T05 mentioned that many handlers teach their guide dogs to find crosswalk buttons, trash cans/dumpsters, mailboxes, poles, and even ATMs through additional training, which serve as important navigational landmarks. This was also evidenced by H03 during the observation session, who used his guide dog to locate specific poles and drain grates while walking in his local community. Additionally, some handlers introduced new vocabulary to address their unique living environments. H21, for instance, taught her guide dog to find exit stairs due to frequent fire alarms in her apartment building. Sometimes, handlers like H23 shared their experiences of teaching new vocabulary to refer to specific destinations: ``When I used to go to my psychology building, [...] I could simply say, `To psychology', and my guide dog knew that we were headed to that specific building.'' Some handlers also taught new commands to demonstrate a specific behavior to fit their needs. For instance, H23 explained that she taught her guide dog a custom cue, “follow”, to enable the dog to trail a particular person at a slower pace, a skill not originally included in her guide dog's training program. This adjustment proved useful for navigating crowded areas. Similarly, H06, H07, H09, H11, H15, and H17 trained their dogs to find hand railings or banisters near stairs to assist their physical weaknesses when ascending or descending stairs. 

Another aspect of personalization that handlers emphasize is the guide dog's stopping criteria during navigation, often refined through trial-and-error experiences. While guide dogs receive foundational training to stop when encountering curbs and potential hazards, they may not initially grasp the handlers' subtle preferences. For instance, H20 prefers stopping at every minor obstacle, such as cracks and tree roots, but mentioned that his guide dog would continue unless it's a curb due to their relatively short time together (i.e., two weeks). Conversely, H14 and H22 preferred to bypass low-rise toe-trip obstacles. Therefore, H22 conducted further training to instruct the dog to halt only when obstacles reached a height of two inches or higher. Trainers noted that these preferences depend on the handler's physical abilities, such as balance and residual vision, as well as the handler's age. 

Interestingly, some handlers selectively bypass designated stops, contrary to guide dog trainers' instructions, for the sake of efficiency. For example, H08 relies on her residual vision and occasionally gives her guide dog a left turn command just before reaching a curb, a practice she acknowledges she shouldn't do but finds more efficient.

\begin{quote}
    {“And I do something that I probably shouldn't do, but at a certain point in my neighborhood, instead of making her walk me all the way to the end of the sidewalk and stop at the curb, I give her a left turn command right before we reach the curb. [...] And I shouldn't do that. I should probably always make her stop at the curb, but it just seems like it's more efficient to just, ‘We're going to turn left. Just go up there. I'll go up to the light pole, turn left and off we go.’”} - \textbf{H08}
\end{quote}

Geographic location and environmental factors also influence where handlers choose to stop. Handlers in colder regions of the United States, like H11 and H21, often encounter broken sidewalks with large cracks. We observed that H21’s guide dog stopped at every crack along the sidewalk, as shown in Figure~\ref{fig:stop}, during the observation session. On the other hand, H18, who resides in a rural area with less distinct curbs, mentioned that her guide dog stops at the edges of roads or ramps. H22 also shared an interesting experience of using her familiarity with her surroundings to personalize the stopping criteria in busy city environments.

\begin{quote}
    {“Now on the New York City subway, a lot of times the stairway is very close to the track and you have to go around these posts and there are people in the way. And so she sees that as unsafe. So there are times I have to encourage her to go along, even if she thinks it's unsafe, which is something they teach you definitely not to do in school. [...] But I grew up in New York, so I know New York. Yeah, I know the New York City system. [...] So they're taught not to go over those, but there are times you do have to walk on those.”} - \textbf{H22}
\end{quote}

Despite the extensive personalized training required for class-ready guide dogs, their working lifespan is relatively short, with retirement typically occurring at around eight to ten years of age. Consequently, handlers find the process of personalization burdensome, as they must repeat it each time a dog retires and they start working with a new one. H20 stated that

\begin{quote}
    {“It's a learning experience for both you and the dog. And of course, just when you and the dog really get good, the dog is getting older and it's almost time to think about retiring the dog.”} - \textbf{H20}
\end{quote}

\subsection{Handlers' Expectations and Concerns Towards Guide Dog Robots}
This section explores key facets related to the expectations and concerns of guide dog handlers in their prospective collaboration with guide dog robots. While we cannot anticipate their perspectives grounded on direct experience with guide dog robots until this technology becomes a reality, interviews with handlers and trainers provide valuable insights into the critical considerations that should inform the development of this emerging technology. In general, our findings indicate a high likelihood of acceptance for this innovative technology, driven by their optimistic expectations of its potential to address various challenges faced by guide dogs. However, the interviewers also highlighted several inherent challenges of artificial systems, such as difficulties in establishing trust and a sense of agency.  

\subsubsection{Expectations}
\label{sec:431-expectations}
During our interviews, we received unequivocally positive feedback about the concept of a guide dog robot. This enthusiasm should be tempered with the consideration that the respondents, given their generally positive disposition toward robotic systems, might be biased in their preferences for technology. Furthermore, the interviewers expressed their opinions with the underlying assumption that the robot would provide navigation services with the same proficiency as animal guide dogs. Nonetheless, a common belief emerged among both handlers and trainers that technologies, including guide dog robots, represent a promising avenue for enhancing their mobility and functionality. This optimism stems from artificial systems’ capacity to expedite development cycles, seamlessly integrate advanced sensing and computing capabilities, and provide the flexibility to perform additional functions, as highlighted by insights from participants H03, H09, and H11.

\begin{quote}
    {“So, I tell people all the time when I speak that technology is a hundred times faster than medicine. To find a cure for my eyes I’m hopeful, but not banking on it. Where I’m banking on [is] technology, just like you guys with the robot.”} - \textbf{H03}
\end{quote} 

During the course of our interviews, we discovered that guide dog handlers hold the belief that guide dog robots have the potential to alleviate critical inherent challenges that handlers typically encounter when working alongside living animals. H05 and H08 highlighted the advantage of a robot by providing examples of BLV individuals who cannot work with guide dogs due to allergies and financial constraints. H03 added a positive outlook of a guide dog robot based on personal experiences with discrimination while traveling to another country having to be quarantined for over a month due to different disability laws and acts. H23 added that the most common issue is ride-share discrimination, despite the Americans with Disabilities Act (ADA) federal law against it,~\bt{a finding that aligns with \ft{the lived experience of one of the authors of this paper and} observations made by Rickly et al.}~\cite{rickly2021}. H18 suggested that career professionals may prefer a robot due to apartment restrictions on pets. H09 pointed out that a robot could assist BLV individuals with limited mobility and orientation skills who cannot have a guide dog. H19, the second oldest subject, mentioned the benefit of physically being able to lean on a robot as her walking and balance abilities decline with age. H08 concluded that a guide dog robot would bridge these gaps:

\begin{quote}
“But that’s why I’m excited about your robot because it’s going to fill a niche that a dog can’t fill. . . . I think what you’re doing is really necessary and I’m really all for it. I think that when I’m 90 years old, a dog is going to be too much for me, but a robot will be perfect.” - \textbf{H08}
\end{quote}

Several handlers viewed a robot as an integrated system that is able to contain multiple sensors. Specifically, H11 anticipated improved vision through advanced cameras, surpassing the capabilities of dogs. H09 sought an integrated GPS system, eliminating the need for using multiple apps as discussed in Section~\ref{subsec:inflexibility}.

\begin{quote}
“So all of that stuff can be put into one box. I'm all for it. [...] But again, it's using all of these different apps. I want it all in one box. And a robot will do that.” -  \textbf{H09}
\end{quote} 

In addition, H03 anticipated functionality enhancements to bolster security:

\begin{quote}
“So I'm thinking for security, you probably should have something like... Every day in America, a blind guy with a cane is hit over the head and they steal his wallet. [...] I think you should have some type of button with an alarm and with some type of 9-1-1 call, so that you could provide a certain amount of security feeling, a safe feeling, for the blind person.” -  \textbf{H03}
\end{quote} 

\bt{Opinions about the robot's appearance vary: some handlers value visibility to the public, while others lean towards a more inconspicuous design. Additionally, there is a spectrum of preferences concerning dog-like attributes, including the presence of a head, ears, and fur.}

\subsubsection{Trust}
Guide dogs’ navigation assistance holds profound importance beyond simply offering convenience for \ft{handlers}. Handlers entrust their very lives to their guide dogs. When BLV individuals begin to work with a guide dog, they commonly do not mainly use white canes, but mainly work with guide dogs. Therefore, the responsibility for mobility and the inherent risks associated with community navigation—such as collisions with cars, falling into holes, or tripping over obstacles—shifts to the dog. Given this heavy reliance on guide dogs for crucial safety, trust between the handler and the dog emerges as a paramount factor to be addressed in the development of guide dog robots. Indeed, some handlers expressed strong concern about the difficulty of establishing trust with a robotic system. H17 stated, “I mean, that trust is, I think, a very important component of the whole thing. And you're not going to get that from a robot.” She further speculated that guide dog handlers might not use robots for such a reason, but people who are not working with a guide dog could adopt them. H05 questioned about building a bond with a guide dog robot, saying:

\begin{quote}
    {“I think something like this robot would be good as far as the right, the left, the down, the guide, but you still don't have that special bond that you do with the dog and person. There's such a relationship.”} - \textbf{H05}
\end{quote}

However, our finding indicates that trust is not merely a byproduct of the inherent nature of animal dogs. H05 highlighted the unique bond between handlers and their guide dogs compared to regular pets: “A lot of people have pets and they're real close to them, but it's not the same as when you're trusting your life to a dog.” The trust developed with a guide dog is rooted in a long-term partnership and shared experiences. In fact, several handlers admitted that they did not trust their guide dogs in the beginning. H18, for instance, mentioned this while explaining her experience with her seventh guide dog, with whom she had been partnered for only nine days at the time of the interview:

\begin{quote}
{“We are all over the place with this new dog. Well, I've only been with her for nine days, so it's still very new and very unpredictable. [...] Well, for one thing, confidence, the longer I've worked with a guide dog, I trust my life to that dog and to my new dog, I don't yet trust my life to this new dog. Not yet, but we're working together.”} - \textbf{H18}
\end{quote}

H18 mentioned that she would do more verbal communication with the new guide dog to establish a bond so that the guide dog can get used to her voice, distinguishing it from others. H03, H10, and H17 noted that trust is cultivated over time as they and their guide dogs navigate challenging situations, ones that handlers could not manage on their own, like safely avoiding obstacles. H15 described a defining moment that solidified his trust in his guide dog during training with his instructor in a bustling area:

\begin{quote}
    {“We turned around and started back, there were lots of people and I started thinking we were going to be hours. They were still under construction with the no end. And I said about all this stuff and all of a sudden I came back to reality and said, ‘This dog is getting through all these people, all these obstacles.’ They always tell you trust your dog. And at that moment I believe it completely. It was very enlightening.”} - \textbf{H15}
\end{quote}

From our interviews, it became evident that trust is rooted in a dynamic interaction that demands a sustained, long-term partnership, personalization, and most importantly, a high level of reliability in the services that should be provided. 

\subsubsection{Sense of Agency}
Sense of agency, denoting a user’s perception of control over actions and their outcomes~\cite{gallagher2000philosophical,moore2016sense}, is a critical aspect that should be considered when creating a system with autonomy~\cite{berberian2012automation}. One noticeable indicator of a high sense of agency over a particular system is a user's willingness to take responsibility for undesirable results~\cite{shneiderman2016designing, haggard2009experience}. We found guide dog handlers’ profound sense of agency while collaborating with their canine companions. Notably, handlers, including H01, H03, and H05, readily took responsibility for their actions and outcomes, even in accident scenarios. H05 shared her experience of an accident, “I've been hit by a drunk driver with my first dog before and it wasn't my dog's fault, it was my fault, partly.” While guide dogs offer a high degree of autonomy, where maintaining a strong sense of agency is generally difficult, handlers expressed high self-responsibility based on clear expectations regarding the dogs’ abilities. Such innate comprehension may be absent in robotic systems, potentially posing a significant barrier to the adoption of robotic guide dogs.

Transitioning from guide dogs to robotic counterparts requires thoughtful attention to user agency. One potential approach is ensuring that handlers retain controllability during navigation, enabling them to intervene as necessary. This sentiment was underscored by handlers, including H19, H21, and H22, who stressed the need to maintain authority over the robot in diverse real-world contexts. Both H21 and H22 drew attention to situations that necessitate on-the-fly adjustments in the robot's navigation. H21 mentioned, “I mean, I think that there are times where it's appropriate to be able to type in something and just go straight there. But I mean, things happen.” H19 further accentuated the significance of control:

\begin{quote}
    {“Easily change the program on the spot. And if you can't do that, it might not be as useful as a dog.”} - \textbf{H19}
\end{quote}

\section{Discussion}
In this study, we aimed to bridge the gap between guide dog robot developers and stakeholders (i.e., guide dog handlers and trainers) by gaining user-driven insights into handler-guide dog interactions based on semi-structured interviews with 23 guide dog handlers and 5 professional guide dog trainers. Thematic analysis revealed key findings crucial for guide dog robot design, encompassing (1) limitations in handler-guide dog interaction, (2) the importance of continuous interactions to fulfill handlers' needs, and (3) handlers' perspectives on guide dog robots. 

This paper offers unique insights into guide dog robot development. While numerous studies have delved into the capabilities and limitations of guide dogs~\cite{miner2001experience,whitmarsh2005benefits,schneider2005practice,wiggett2008experience,craigon2017she,lloyd2008guide}, and the interaction between assistive devices and \ft{BLV individuals}~\cite{ulrich2001guidecane, slade2021multimodal, guerreiro2019cabot, zhang2023follower, kuribayashi2023pathfinder}, efforts to understand and analyze users' requirements with respect to mobile robots remain sparse. Many studies, even those proposing mobile assistive devices, have either primarily emphasized autonomous navigation in new environments\bt{ ~\cite{guerreiro2019cabot, meldog, kuribayashi2023pathfinder}} or overlooked users' perspectives in design \cite{xiao2021robotic,chen2023quadruped, hamed2019hierarchical, defazio2023seeing, due2023guide}. Thus, the intricate dynamics between guide dogs and handlers in their daily lives, particularly where handlers take an orientation role, remain under-explored. This paper aims to bridge this knowledge gap, drawing on the real-world experiences of stakeholders. In our discussion, we consolidate key aspects of robot development, integrating our findings with existing research, and suggest design considerations that encompass the hardware, perception, and interface facets of a guide dog robot. Additionally, we acknowledge the limitations of our study and outline directions for future research.

\subsection{Design Implications}

\subsubsection{Form Factor}
When designing a guide dog robot, several key factors must be considered, including leg length, overall dimensions, weight, protective features, and appearance. \bt{Regarding the robot's size, it should be compact enough to fit beneath seats on subways or airplanes, the spaces where guide dogs typically rest, as our findings support that handlers frequently use public transportation.} Simultaneously, it must be large enough to walk over stairs, which typically range from $15$-$19~\si{\cm}$ in height\bt{~\cite{stairs}}. The robot's weight should strike a balance between ensuring portability for an individual and providing enough heft to convey its movements and effectively halt the handler when necessary.

Regarding the guide dog robot's exterior, it should feature adequate waterproofing and dustproofing to protect its internal electronic components \bt{considering that the handler may encounter unexpected rain}. An Ingress Protection (IP) rating of at least IP65 would be appropriate, denoting complete dust resistance and protection against low-pressure water jets from all directions. Thermal management should also be considered in design to operate in extreme weather~\cite{Tucker84}. \bt{In terms of appearance, as reported in Section~\ref{sec:431-expectations}, we may personalize it according to the users' needs.}

\subsubsection{Sensor\bt{s and Security}}
We identified important considerations for the sensor suite of guide dog robots. One pivotal aspect is mitigating the vision sensors' blind spots. Current guide dog robots predominantly utilize vision sensors, incorporating a combination of stereo cameras~\cite{hwang2023system}, depth sensors~\cite{xiao2021robotic, chen2023quadruped, kim2023train, kim2023transforming}, and LiDAR sensors~\cite{xiao2021robotic, chen2023quadruped, kim2023train, kim2023transforming, defazio2023seeing}. However, an exclusive dependence on vision can be problematic. There are inherent blind spots made by the handler standing adjacent to the robot, and potential inaccuracies due to close proximity sensor measurements, as highlighted by Liu et al.~\cite{liu2023designing}. While guide dogs utilize their vision, they also employ their acute sense of hearing to detect approaching objects~\cite{longestablishing}. Thus, diversifying sensor modalities, perhaps by integrating acoustic sensing, could amplify the perceptual capabilities of guide dog robots, ensuring safer navigation.

\bt{Addressing the detection of overhead obstacles, a known limitation \ft{of} guide dogs~\cite{williams2013pray,jeamwatthanachai2019indoor}, is pivotal in developing guide dog robots. To mitigate this \ft{problem}, the integration of overhead monitoring sensors is essential. Promising solutions include the use of stereo cameras, strategically positioned on the robot as demonstrated in CaBot~\cite{guerreiro2019cabot}, and gimbal-mounted cameras, similar to the approaches in Xiao et al.~\cite{xiao2021robotic} and Chen et al.~\cite{chen2023quadruped}. These technological adaptations may enhance a critical area of safety, addressing issues identified with animal guide dogs.}

\bt{In the context of enhancing security and ensuring that robotic operations are limited to authorized users, exploring a variety of authentication methods is pivotal. For example, haptic or biometric authentications present a viable solution, enabling the identification of registered users with the added benefit of customizable privacy options~\cite{eick2020enhancing}. It is essential to consider diverse authentication strategies for reducing privacy vulnerabilities in robot design.}

\subsubsection{Battery}
Our findings suggest that the robot's battery should sustain operations for more than two hours. This recommendation stems from the average walking habits of our participating guide dog handlers, most of whom primarily depend on walking or public transport. Notably, commercial quadruped robots from Unitree Robotics, advertise a battery life of 2 to 4 hours~\cite{ur23}. However, designers should consider that this duration could diminish due to various factors \bt{such as extra sensors (e.g., LiDAR) and the energy to pull a user ($\sim 40~\si{\newton}$~\cite{hwang2023system})}.

\subsubsection{Perceptual Intelligence}
Recognizing objects for navigation is an indispensable technology for guide dog robots, serving as the handlers' ``eyes''. Although computer vision has advanced significantly with the advent of deep learning techniques~\cite{lecun2015deep}, focused research in the context of guide dog robots remains scarce. Unlike road environments, which predominantly feature distinct lanes and obstacles with limited variations such as vehicles and pedestrians, sidewalks introduce a myriad of unique challenges. A guide dog robot must adeptly navigate along sidewalks, identifying a safe trajectory while recognizing diverse obstacles, from electric scooters and bicycles to overhead tree branches and sidewalk cracks. Notably, while there are expansive datasets tailored for autonomous vehicles~\cite{cordts2016cityscapes, yu2020bdd100k, sun2020scalability, geyer2020a2d2, caesar2020nuscenes}, they \bt{may} not encapsulate the specificities of environments and sensing requirements intrinsic to guide dog robots.

To bridge this gap, Park et al.~\cite{park2020sideguide} pioneered the development of a dataset from cameras oriented from a pedestrian's viewpoint. \bt{More recent work~\cite{karnan2022socially, zhang2023towards} entailed large-scale multimodal sensor data collected from a university campus environment using mobile robots. Integrating such data with various learning methods, such as learning-from-demonstration~\cite{ravichandar2020recent} and visual navigation methods that employ self-supervised learning~\cite{shah2021ving,karnan2023sterling} and imitation learning~\cite{sorokin2022learning,ai2022deep}, could enable adept interpretation of contextual cues for decision-making in robots. Sorokin et al.~\cite{sorokin2022learning} proposed a learning framework that uses knowledge learned from simulation data and performs trajectory following with a quadruped robot that avoids obstacles and remains on sidewalks for $3.2~\si{\km}$. Shah et al.~\cite{shah2021ving} similarly showed adaptability to a novel environment, although requiring an hour of demonstration. More recently, with the advent of Transformers~\cite{vaswani2017attention, dosovitskiy2020image}, there has been research enabling zero-shot transfer for various computer vision tasks, such as classification~\cite{radford2021learning} and segmentation~\cite{kirillov2023segment}, opening possibilities for perception models to adapt and respond appropriately in an unseen environment. Notable progress includes enabling zero-shot object goal navigation in simulated environments~\cite{zhao2023zero} and achieving long-horizon navigation without any fine-tuning in the target environments~\cite{shah2023lm}.}

\bt{However, these developments are primarily confined to simulations or constrained systems, highlighting that extending these capabilities to more generalized, real-world navigation scenarios continues to be an active and challenging area of research. Therefore, while progress in this field is encouraging~\cite{sridhar2023nomad}, the capability for autonomous navigation in diverse and unpredictable environments remains an active area of research and development. Also, although there have been significant advancements in perception models~\cite{Jocher_YOLO_by_Ultralytics_2023, wang2023image, chen2018encoder}, safety-guaranteed real-world deployment \ft{has} not been accomplished yet. Common failures of object detection such as missing objects (false negative) or detection of non-existent objects (false positive), could potentially lead to hazardous situations (e.g., stopping in the middle of a street or failing to stop at curbs). Prior research has explored using additional information through sensor fusion and considering object properties and context to reduce misclassifications~\cite{man2023person}. In addition to the effort for enhancing algorithm accuracy, leveraging multimodal data (e.g., visual and auditory) or human-in-the-loop systems would alleviate the issues in perception to guarantee safe and reliable decision-making.} 

\bt{As our findings in Section~\ref{subsec:unidirectional} suggest, using natural language for scene descriptions can enhance the safe navigation of BLV people. Recent advancements in large multimodal models~\cite{liu2023visual,zhu2023minigpt,chen2023minigpt} that combine vision encoders with large language models (LLMs) such as GPT-4, as well as open-source LLMs~\cite{touvron2023llama,vicuna2023}, have enabled natural language scene descriptions at a level comparable to human-level artificial general intelligence~\cite{bubeck2023sparks}. Specifically, these large multimodal models process natural language prompts and visual information to generate appropriate natural language responses. This capability allows them to address queries such as “Is it safe to cross the street?” or “Is this seat empty?” \ft{These technologies} are already being applied in vision-assistive mobile applications, including OKO~\cite{oko} and Be My Eyes~\cite{bme}. Future research could productively focus on tailoring perceptual models to deliver valuable information to users~\cite{hoogsteen2022beyond} and integrating these systems into guide dog robot applications.}

\subsubsection{Communication Method}
To address the limitations of unidirectional communication from a handler to a guide dog, guide dog robots need to relay pertinent information to users when necessary. Audio feedback has been widely used for this purpose~\cite{gleason2018footnotes, liu2023designing, liu2023dragon, guerreiro2019cabot}. Yet, from our interviews, we learned the importance of moderating the amount of audio feedback. H09 illustrated this concern, sharing his experience with an electronic travel aid in an urban environment. He noted that excessive information can distract individuals who are relying on their remaining senses, which also agrees with the prior study on cognitive load~\cite{paneels2013listen}. 

In terms of communication from handlers to robots, there exists merit in retaining traditional methods such as verbal cues and hand gestures due to their proven efficacy over time. Audio interfaces for command input, as explored in previous research~\cite{liu2023dragon, liu2023designing, kalpana2020voice, kim2023transforming}, present a natural and effective way to convey the handler's intentions. \ft{
However, their efficiency may wane in loud environments where ambient sounds can obscure commands, when the handler experiences vocal issues like a cold, or when conveying sensitive commands, such as ``find a restroom'', is not desirable.} Given the adaptability of robotic systems, considering alternative modes of communication, such as haptic devices or button interfaces~\cite{hwang2023system}, could offer more diverse ways to interact.

\subsection{Towards Real-World Deployment Beyond Development}
To fill in the gap between the successful integration of the aforementioned guide dog design implication\ft{s} and real-world deployment, we need to focus on establishing a strong bond between handlers and guide dog robots, emphasizing two pivotal factors--- trust and personalization---tailored to handers’ preference that account for their physical attributes and living environments.

We contend that trust building is a \ft{multifaceted} process involving the reliability of the system and long-term partnership. For instance, we found that trust in the handler-guide dog partnership grows over \ft{time and} experience through trial and error. We additionally argue that instilling a sense of agency~\cite{moore2016sense, gallagher2000philosophical,gallagher2012multiple} in users towards the robot is crucial to cultivate trust, a concern frequently raised regarding guide dog robots among the handlers. To \ft{establish trust}, fully autonomous operation of guide dog robots, when paired with users, may not be ideal~\cite{zhang2023follower}, especially during the initial interaction stages where trust bias may arise~\cite{freedy2007measurement} and unexpected robot behavior can diminish trust~\cite{lyons2023explanations}. Similar to novice guide dog handlers who initially lack trust in their guide dogs but gradually develop strong bonds over time, we envision a parallel trajectory for guide dog robots. As trust between handlers and their guide dogs continues to develop \ft{and solidify}, handlers become more comfortable granting their guide dogs greater autonomy, even in the presence of limited communication and interactions between them, and \ft{of the} occasional accidents that may occur during the trust-building process. By mirroring this pattern in the human-robot relationship, we open the door to opportunities for guide dog robots to assume \ft{increasing} autonomous navigation responsibilities as the bond between the user and the robot matures. This approach recognizes that trust is not an instantaneous outcome but a dynamic process that grows over time with shared experiences and a deepening sense of reliability. Therefore, as guide dog robots become more familiar and integrated into the user's life, they can gradually assume greater autonomy in navigation tasks.

Personalization is essential to accommodate the diverse preferences of users with varying physical attributes and living environments~\cite{pineau2003towards, broadbent2009acceptance}. Addressing these varying needs will pose a challenge for a one-size-fits-all decision-making algorithm. For instance, the preset stop criteria for toe-trip obstacles may incur safety issues or inconvenience for users with limited balance who prefer the robot to halt at every toe-trip obstacle or others who have more agility and prefer to pass \bt{minor} toe-trip obstacles, respectively. Algorithms following the established practice (e.g., advancing to the curb's end before turning) may prove less efficient for individuals with remaining visual \ft{function}, as they possess a general sense of when and where to make the turn before reaching the curb. Conversely, having the algorithm turn before reaching the curb may pose safety concerns and impede orientation for individuals who are totally blind. One potential solution to fulfill such a wide range of personalization needs would be to utilize lifelong (or continual) learning techniques~\cite{parisi2019continual}. This adaptive approach ensures that the guide dog robot evolves to align with each user's distinct requirements, ultimately enhancing the overall user experience and the effectiveness of the robot as a mobility aid.

\subsection{Limitations and Future Work}
Our research efforts, while promising, come with inherent limitations that deserve acknowledgment. \bt{It is important to note the potential influence of participant bias in our findings, given the diversity within our participant pool concerning vision levels, living areas, age, and familiarity with technology.} \bt{The majority of participants were recruited from the United States except for one participant from the Republic of Korea. However, we did not observe significant variances in the handler-guide dog interactions between the two countries. Further research may be beneficial for a comprehensive understanding of the nuances of handler-guide dog interactions across different countries.} \ft{Furthermore, the discussions were drawn from interviews exclusively with guide dog handlers and trainers who lack experience with guide dog robots. Engaging guide dog robot developers and engineers in future research could yield more comprehensive insights into the technological feasibility and progression towards the envisioned goals of guide dog robots.}

In light of these limitations, our research paves the way for several directions of future work aimed at addressing the identified \bt{challenges}, validating our findings through user studies, exploring personalization, \bt{investigating companionship between the handler-guide dog unit,} and contributing to the evolving field of trust in human-robot interaction. First, it is essential to integrate the design implications derived from our current research into the development of our guide dog robot system. This will involve refining the robot's capabilities to ensure it can reliably guide the user on a sidewalk having multiple obstacles including overhead obstacles and co-designing the harness handle interface. This foundational step will set the stage for more advanced interactions between the handler and the guide dog robot during indoor and outdoor navigation tasks \ft{in more complex environments}. Second, it remains crucial to establish objective measures of the robot's performance and its impact on users' daily lives by conducting both controlled and in situ user studies with BLV individuals. These future studies will enable us to gather measurements to evaluate our guide dog robot and empirical data on trust and usability in real-world scenarios. Third, we recognized the importance of personalization in accommodating the unique needs and preferences of guide dog handlers. To better understand these important factors, it is imperative to conduct interviews with a diverse range of participants, such as individuals having different vision levels, using different types of mobility aids, and having additional disabilities, in order to explore the nuances of personalization and tailor our system accordingly. \bt{Fourth, our study primarily delved into interaction dynamics related specifically to navigational tasks, leaving a gap in our investigation regarding the companionship and emotional support aspects lacking in guide dog robots. The benefits of companionship and emotional support in the handler-animal guide dog relationship are well-documented~\cite{whitmarsh2005benefits, wiggett2008experience}, a finding also evident in our interviews. As future users of guide dog robots may have the choice to adopt a dog as a pet or customize the guide dog robot to offer companionship and emotional support~\cite{baecker2020emotional, andreasson2018affective, guo2023touch}, future research in these areas seems pivotal.} Finally, an important avenue for future research lies in the broader themes related to human-robot teaming and trust, as highlighted in previous studies~\cite{hancock2011meta,lewis2018role,billings2012human}. These explorations will collectively contribute to our understanding of the dynamics of human-robot teams and the development of an effective guide dog robot.

\section{Conclusion}
In this paper, we propose a design implication for legged guide dog robots that are actively being developed. Data from 23 guide dog handlers and 5 trainers revealed limitations of the handler-guide dog interaction, personalization aspects that are beyond the official guide dog training regimen, and multifaceted perspectives surrounding guide dog robots. These findings provide important new insights and strongly advocate for the importance of our user-driven design approach. Moreover, our findings emphasize the significance of trust and personalization in achieving real-world adoption, which is critical for future guide dog robot development. Our study contributes valuable insights to the HCI and robotics communities, advancing the prospect of an additional mobility option tailored to the needs of BLV individuals.


\begin{acks}
    \ft{The authors originally created the text and content, while ChatGPT was used exclusively for refining the writing and checking the grammar.}
\end{acks}


\bibliographystyle{ACM-Reference-Format}
\bibliography{sample-base}

\appendix

\clearpage
\section{Guide dog evaluation}
\label{sec:evaluation}

\ft{Guide dogs are comprehensively evaluated to become "class-ready" guide dogs. We have categorized the evaluation components acquired from the Guide Dog Users, Inc. (GDUI) School Survey Summary and presented them in Table~\ref{tab:evaluation}.}

\begin{table*}[ht!]
  \caption{Final testing components - Guide dog evaluation metric}
  \label{tab:evaluation}
  \begin{tabular}{ll}
    \toprule
    \textbf{Task title} & \textbf{Evaluation components} \\
    \midrule
    Obedience & Respond to basic obedience commands\\
     & Reliably come when called while off-leash\\
    \midrule
    Guide work & Work in a variety of street crossing settings \\
     & Work in a variety of street crossing settings \\
     & Trained in urban, suburbia (small towns), and rural environments \\
     & Reliably stop at changes in elevation, i.e., steps, curbs, and drop-offs \\
     & Reliably avoid or indicate obstacles \\
     & Reliably avoid or indicate overhead obstacles \\
     & Reliably turn left, right, or move forward on the hand signal from the handler \\
     & Reliably turn left, right, or move forward on the verbal command of  the handler
 \\
     & Enter, exit, and ride quietly in passenger cars or other vehicles \\
     & Enter, exit, and ride quietly on buses \\
     & \makecell[l]{Enter, exit, and ride quietly on subways, light rail systems, or passenger trains \\ \quad and safely navigate elevated platforms} \\
     & Exposed to security checkpoints such as those found in airports \\
    \midrule
    Buildings & Ride escalators \\
    & Safely navigate revolving doors with their handler \\
    & \makecell[l]{Work in environments such as university campuses or shopping malls \\ \quad where there are multiple locations that must be accessed via nonlinear pathways} \\
    \midrule
    Traffic & Work safely in traffic \\
    & Safely negotiate “traffic checks” \\
    & Disobey the handler's commands when it would be unsafe to comply (intelligent disobedience)\\
    \bottomrule
\end{tabular}
\Description{Table 3: "(Table is under Appendix A - Guide Dog Evaluation) Final testing components - Guide dog evaluation metric. This 5-by-2 table contains information for the evaluation components that a ‘class-ready’ guide dog completed. Each row of the first column represents the task title, including ‘Obedience’, ‘Guide work’, ‘Buildings’, and ‘Traffic’. The second column contains the description of the evaluation components for each task.”}
\end{table*}

\section{Guide Dog Users, Inc. (GDUI) School Survey Summary}
\label{sec:apx-gduisurvey}

\ft{The Guide Dog Users, Inc. (GDUI) School Survey Summary contains information about the expected range of services provided by guide dogs that have completed formal training at various guide dog schools. We compiled the data which is displayed in Table~\ref{tab:gduisurvey}.}

\begin{landscape}
\begin{table*}
  \caption{Guide dog's assured ability upon completion of formal training}
  \label{tab:gduisurvey}
  \small
  \begin{tabular}{lccccccccccccc}
  %
    \toprule
    Q: Upon completion of formal training, clients can be assured that a dog from your program & A & B & C & D & E & F & G & H & I & J & K & L\\
    \midrule
    \textit{will be house broken} & \cmark & \cmark & \cmark & \cmark & \cmark & \cmark & \cmark & \cmark & \cmark & \cmark & \cmark & \cmark\\
    \textit{will have been trained to behave well in public, e.g. no barking or stealing food} & \cmark & \cmark & \cmark & \cmark & \cmark & \cmark & \cmark & \cmark & \cmark & \cmark & \cmark & \cmark\\
    \textit{will exhibit no aggression toward people or other animals} & \cmark & \cmark & \cmark & \cmark & \cmark & \cmark & \cmark & \cmark & \cmark & \cmark & \cmark & \cmark\\
    \textit{will respond to basic obedience commands} & \cmark & \cmark & \cmark & \cmark & \cmark & \cmark & \xmark & \cmark & \cmark & \cmark & \cmark & \cmark\\
    \textit{will disobey the handler's commands when it would be unsafe to comply (intelligent disobedience)} & \cmark & \cmark & \cmark & \cmark & \cmark & \xmark & \cmark & \cmark & \cmark & \cmark & \cmark & \cmark\\
    \textit{will be trained to work safely in traffic} & \cmark & \cmark & \cmark & \cmark & \cmark & \cmark & \cmark & \cmark & \cmark & \cmark & \cmark & \cmark\\
    \textit{will be trained to safely negotiate “traffic checks”} & \cmark & \cmark & \cmark & \cmark & \cmark & \cmark & \cmark & \cmark & \cmark & \cmark & \cmark & \cmark\\
    \textit{will be trained to work in a variety of street crossing settings} & \cmark & \cmark & \cmark & \cmark & \cmark & \cmark & \cmark & \cmark & \cmark & \cmark & \cmark & \xmark\\
    \textit{will be trained to reliably stop at changes in elevation, i.e. steps, curbs and drop-offs} & \cmark & \cmark & \cmark & \cmark & \cmark & \cmark & \cmark & \cmark & \cmark & \cmark & \cmark & \cmark\\
    \textit{will be trained to reliably avoid or indicate obstacles} & \cmark & \cmark & \cmark & \cmark & \cmark & \cmark & \cmark & \cmark & \cmark & \cmark & \cmark & \cmark\\
    \textit{will be trained to reliably avoid or indicate overhead obstacles} & \cmark & \cmark & \cmark & \cmark & \cmark & \cmark & \cmark & \cmark & \cmark & \cmark & \cmark & \cmark\\
    \textit{will be trained to reliably turn left, right or to move forward on the hand signal from the handler} & \cmark & \cmark & \cmark & \xmark & \cmark & \cmark & \cmark & \cmark & \cmark & \cmark & \cmark & \cmark\\
    \textit{will be trained to reliably turn left, right or to move forward on the verbal command of the handler} & \cmark & \xmark & \cmark & \cmark & \cmark & \cmark & \cmark & \cmark & \cmark & \cmark & \cmark & \cmark\\
    \textit{will be trained to ride escalators} & \cmark & \cmark & \cmark & \cmark & \cmark & \cmark & \xmark & \cmark & \cmark & \cmark & \cmark & \cmark\\
    \textit{will be trained to follow members of the public when commanded to do so by the handler} & \xmark & \cmark & \cmark & \cmark & \cmark & \cmark & \xmark & \cmark & \cmark & \cmark & \cmark & \cmark\\
    \textit{will be trained to enter, exit and ride quietly in passenger cars or other vehicles} & \cmark & \cmark & \cmark & \xmark & \cmark & \cmark & \cmark & \cmark & \cmark & \xmark & \cmark & \cmark\\
    \textit{will be trained to enter, exit and ride quietly on buses} & \cmark & \cmark & \cmark & \cmark & \cmark & \cmark & \cmark & \cmark & \cmark & \cmark & \cmark & \cmark\\
    \makecell[l]{\textit{will be trained to enter, exit and ride quietly on light rail systems or passenger trains} \\ \quad \textit{and to safely navigate elevated platforms}} & \cmark & \cmark & \cmark & \xmark & \cmark & \cmark & \cmark & \cmark & \cmark & \xmark & \cmark & \cmark\\
    \textit{will have been exposed to security checkpoints such as those found in airports} & \cmark & \cmark & \cmark & \cmark & \cmark & \xmark & \cmark & \cmark & \cmark & \xmark & \cmark & \cmark\\
    \textit{will have been trained in urban environments} & \cmark & \cmark & \cmark & \cmark & \cmark & \cmark & \cmark & \cmark & \cmark & \cmark & \cmark & \cmark\\
    \textit{will have been trained in small towns or suburbia} & \cmark & \cmark & \cmark & \cmark & \cmark & \cmark & \cmark & \cmark & \cmark & \cmark & \cmark & \cmark\\
    \textit{will have been trained in rural environments} & \cmark & \cmark & \cmark & \cmark & \cmark & \cmark & \cmark & \cmark & \cmark & \cmark & \cmark & \cmark\\
    \textit{will have been trained to guide safely along roads with no sidewalks or footpaths} & \cmark & \cmark & \cmark & \cmark & \cmark & \cmark & \cmark & \cmark & \cmark & \cmark & \cmark & \cmark\\
\makecell[l]{\textit{will have been trained to work in environments such as university campuses or shopping malls} \\ \quad \textit{where there are multiple locations which must be accessed via nonlinear pathways}} & \cmark & \cmark & \cmark & \cmark & \cmark & \cmark & \cmark & \cmark & \cmark & \cmark & \cmark & \cmark\\
    \textit{will be trained to reliably retrieve dropped items for the handler} & \xmark & \xmark & \cmark & \xmark & \xmark & \xmark & \xmark & \xmark & \cmark & \xmark & \cmark & \xmark\\
    \textit{will be trained to locate some specific objects/landmarks on command} & \cmark & \cmark & \cmark & \xmark & \cmark & \cmark & \cmark & \cmark & \cmark & \cmark & \cmark & \cmark\\
    \textit{will be trained to reliably come when called while off leash} & \cmark & \cmark & \cmark & \cmark & \cmark & \cmark & \cmark & \cmark & \cmark & \cmark & \cmark & \cmark\\
    \textit{will be trained to formally guide on leash} & \xmark & \xmark & \xmark & \xmark & \cmark & \xmark & \xmark & \xmark & \cmark & \xmark & \cmark & \xmark\\
    \bottomrule
\end{tabular}
\Description{Table 4: "(Table is under Appendix B - Guide Dog Users, Inc. (GDUI) School Survey Summary) Guide dog’s assured ability upon completion of formal training. This table summarizes the abilities of what clients may expect from guide dogs trained from 12 guide dog schools. The first column contains the expected abilities of a guide dog and each column from the second column corresponds to a unique guide dog school containing a ‘check’ or ‘x’ mark for each row that indicates the expectation.”}
\footnotesize{\\Order of the organizations is random.}\\
\end{table*}
\end{landscape}

\end{document}